\documentclass{article}

\usepackage[preprint]{diffpogan}

\usepackage[utf8]{inputenc} 
\usepackage[T1]{fontenc}    
\usepackage{hyperref}       
\usepackage{url}            
\usepackage{booktabs}       
\usepackage{amsfonts}       
\usepackage{nicefrac}       
\usepackage{microtype}      
\usepackage{xcolor}         
\usepackage{algorithm}
\usepackage{algorithmic}
\usepackage{multirow}
\usepackage{multirow}
\usepackage{makecell}
\usepackage{tabularx}
\usepackage{caption}
\usepackage{textcomp}
\usepackage{xcolor}
\usepackage{graphics} 
\usepackage{epsfig} 
\usepackage{mathptmx} 
\usepackage{times} 
\usepackage{amsmath} 
\usepackage{amssymb}  
\usepackage{bm}
\usepackage{mathrsfs}
\usepackage{wrapfig}

\title{DiffPoGAN: Diffusion Policies with Generative Adversarial Networks for Offline \\ Reinforcement Learning}

\author{%
  Xuemin Hu$^{1}$,  Shen Li$^{1}$,  Yingfen Xu$^{1}$, 
 Bo Tang$^{2}$,  Long Chen$^{3}$\\
  $^{1}$ Hubei University, China, $^{2}$Worcester Polytechnic Institute, USA, \\$^{3}$Chinese Academy of Sciences, China
}

\begin{document}

\maketitle

\begin{abstract}
Offline reinforcement learning (RL) can learn optimal policies from pre-collected offline datasets without interacting with the environment, but the sampled actions of the agent cannot often cover the action distribution under a given state, resulting in the extrapolation error issue. Recent works address this issue by employing generative adversarial networks (GANs). However, these methods often suffer from insufficient constraints on policy exploration and inaccurate representation of behavior policies. Moreover, the generator in GANs fails in fooling the discriminator while maximizing the expected returns of a policy. Inspired by the diffusion, a generative model with powerful feature expressiveness, we propose a new offline RL method named Diffusion Policies with Generative Adversarial Networks (DiffPoGAN). In this approach, the diffusion serves as the policy generator to generate diverse distributions of actions, and a regularization method based on maximum likelihood estimation (MLE) is developed to generate data that approximate the distribution of behavior policies. Besides, we introduce an additional regularization term based on the discriminator output to effectively constrain policy exploration for policy improvement. Comprehensive experiments are conducted on the datasets for deep data-driven reinforcement learning (D4RL), and experimental results show that DiffPoGAN outperforms state-of-the-art methods in offline RL.

\end{abstract}

\section{Introduction}

Offline reinforcement learning attracts researchers' attention due to the advantage of learning from pre-collected datasets without interacting with the environment, which can effectively decrease the risk and cost of training an agent in the fields of autonomous driving \cite{hu2023learning,Hu2024How}, medical treatment \cite{liu2020reinforcement}, and robotics \cite{kober2013reinforcement,agand2023deep}, etc. However, unlike imitation learning \cite{hu2021learning} with a large number of high-quality labeled data and traditional online RL with lots of environmental interactions, the sampled actions of the agent in offline RL cannot often cover the action distribution under a given state, resulting in the overestimation issue for the out-of-distribution (OOD) actions that is also named the extrapolation error problem. In this case, offline RL algorithms need to generate accurate estimation for OOD actions and learn optimal policies that are better than  behavior policies in the dataset.

Classic policy optimization methods often learn OOD actions when they are applied in offline RL, and these learned OOD actions cannot be accurately estimated by the policy evaluation model, thus these methods cannot yet solve the extrapolation error problem. In recent years, deep generation models such as generative adversarial network and diffusion have achieved remarkable success in dealing with high dimensional data such as computer vision (CV) \cite{zhu2017unpaired} and natural language processing (NLP) \cite{zou2023diffusion}, etc. CV and NLP methods aim to generate images or text based on contextual information, while offline RL methods are designed to generate actions or trajectories according to the agent states. Therefore, CV, NLP, and offline RL methods have the same goal of learning from offline datasets, which means that deep generative models can provide an effective approach for learning policies in offline RL. Some methods such as batch-constrained Q-learning (BCQ) \cite{fujimoto2019off} and safe generative batch-constrained Q-learning (SGBCQ) \cite{dong2023safe} apply deep generative models in model-free RL, where generative networks are directly used to generate policies and regularization is used to limit the deviation between learned policies and behavior policies in order to obtain the policies that approximate behavior policies. However, these methods often suffer from insufficient constraints on policy exploration and inaccurate expressiveness of behavior policies, which result in learning sub-optimal policies. 

Generative adversarial networks can theoretically provide a good regularization method for constraining policy exploration, where the discriminator outputs the probability of quantized distribution shift. Unfortunately, it is difficult to apply this method in practice because the generator suffers from simultaneously achieving the two unbalanced goals, fooling the discriminator and maximizing the expected returns, which will cause training instability and make the learned policies mismatch the behavior policies. Although adding auxiliary networks can alleviate this issue, it is a difficult task to design effective auxiliary networks, and this will also significantly increase the network complexity. Recent researches \cite{wang2022diffusion,chen2022offline} have proved that introducing diffusion into offline RL can greatly improve the expressiveness of behavior policies, but it have limited effect on constraining policy exploration, especially for the tasks with sparse reward terms. 

In this paper, we propose a novel method named Diffusion Policies with Generative Adversarial Networks (DiffPoGAN\footnote{Code available at https://anonymous.4open.science/r/DiffPoGAN-BECE.}) for offline reinforcement learning. In the proposed DiffPoGAN, the diffusion is served as the policy generator to generate a diverse distribution, and a regularization method based on the maximum likelihood estimation is developed to encourage the generator to generate data that approximate the distribution of behavior policies, which are joined to make the generator fool the discriminator while maximizing the  expected returns. Besides, we develop an additional regularization term based on the discriminator output to effectively constrain policy exploration for policy improvement since the discriminator in GANs can be used to evaluate the probability that the generated action is in behavior policies. In conclusion, this paper provides following three contributions. 

\begin{itemize}
    \item  A new offline RL method named diffusion policies with generative adversarial networks is proposed to handle the issue that the generator cannot  fool the discriminator while maximizing the expected returns in GAN-based offline RL methods.
    \item We develop an additional regularization term based on the discriminator output to appropriately constrain policy exploration for policy improvement.
    \item We conducted extend experiments on the D4RL, and experimental results demonstrate the proposed method outperforms state-of-the-art methods on the benchmark.
\end{itemize}

\section{Related Work}
\subsection{Offline reinforcement learning}
Extrapolation error is an essential and challenging problem in offline RL. Previous methods for addressing this problem are usually divided into following categories: 1) policy regularization, which regularizes the deviation between the learned policy and behavior policy \cite{fujimoto2019off,fujimoto2021minimalist,wang2022diffusion}, 2) uncertainty estimation, which reduces the model uncertainty or selects the data with low uncertainty to ensure overall stability \cite{wu2021uncertainty}, 3) value function constraint, which assigns low values to OOD data to achieve pessimistic estimation \cite{kumar2020conservative,kostrikov2021offlinefisher}, and 4) considering offline RL as a sequence prediction issue with return guidance \cite{janner2022planning}. In these methods, policy regularization is the most widely used one in offline RL due to its simplicity and the property of ensuring that the Q-value is not overestimated. The proposed method in this paper also belongs to this category. Unlike existing policy regularization approaches that mainly regularize policies by approximating the deviation degree, our method regularizes policies via both the diffusion loss derived from MLE and the discriminator output in the GAN.

\subsection{Generative Adversarial Networks in offline RL}
Previous approaches \cite{fu2017learning,chen2023hierarchical} apply GANs to learn a dynamics model in offline RL. Some recent methods also directly utilize GANs for policy learning in the actor-critic framework \cite{dong2023safe,vuong2022dasco}. However, these methods often suffer from insufficient diversity of the generator to fool the discriminator while maximizing the  expected returns. Vuong et al. \cite{vuong2022dasco} propose the dual-generator adversarial support constrained offline RL method (DASCO) with dual generators to remove the tension between maximizing return and matching the data distribution by generating the mixed data distribution. However, the mixed distribution of the dual generators potentially has long tails and multiple peaks, which will limit the model in a sub-optimal policy. Additionally, applying dual generators introduces larger complexity and results in higher uncertainty. Different from previous methods, our method uses the diffusion as the generator and introduces the diffusion loss derived from MLE to assist the generator in the training process, which can strengthen the distribution match between the generated policy and behavior policy, and effectively handle the issue that the generator fails in fooling the discriminator while maximizing the expected returns.

\subsection{Diffusion models in offline RL}
Generative models have been introduced to enhance the generative capacity of the policy network in previous work. Recently, diffusion models have gained considerable attention in offline RL due to their impressive generative capacities and been used to model trajectories \cite{janner2022planning}, expand datasets \cite{chen2023genaug,lu2024synthetic}, and generate action distributions \cite{wang2022diffusion,he2023diffcps,hansen2023idql}, etc. The most similar work to our method is the Diffusion Q-learning \cite{wang2022diffusion}, which utilizes the diffusion as the actor to generate policies with high expressive capacity, and then regularizes the policy to strengthen the distribution constraint. However, it directly uses the diffusion loss for regularization to constrain policy exploration, thus fails in efficiently learning policies that satisfy the potential data constraint in behavior policies and performing well in some tasks with sparse rewards. Unlike Diffusion Q-learning, we introduce the GAN and employ the discriminator output to evaluate the generated action, which facilitate a reasonable match between the diffusion policy and behavior policy.

\section{Preliminaries} \label{pre}
\textbf{Offline reinforcement learning.} Reinforcement learning is often described as a Markov decision process (MDP) with the tuple $M = \left \{ S, A, P, R, \gamma \right \}$, where $S\in \mathbb{R}^{N}$, $A\in \mathbb{R}^{M}$ , $P \colon S \times A \to S$, $R : S \times A \to \mathbb{R}$, and $\gamma \in ( 0 , 1 ] $ represent the state space, action space, transition dynamics, reward function, and discount factor, respectively. The behavior of an agent in RL is determined by the policy $\pi$, and the goal in RL is to learn an optimal policy $\pi(a|s)$ to maximize the expected discounted return $\mathbb{ E }_{ \pi } [ \sum _{ t = 0 } ^{ \infty } \gamma ^{ t } r ( s _{ t } , a _{ t } ) ]$, where $s_t$ and $a_t$ are the state and action at the time step $t$, respectively. Offline RL algorithms learn a policy from an offline dataset $\mathcal{D}\triangleq\{(s,a,r,s^{\prime})\}$, which is collected by the unknown behavior policy $\pi_{\beta}$ \cite{fujimoto2019off}. $s$, $a$, $r$, and $s'$ represent a state, the action performed on the state, the obtained reward from this action, and the next state after performing this action, respectively.

Offline RL algorithms frequently use the typical Q-learning actor-critic framework for continuous control. During the policy evaluation process, a parameterized critic network, which is usually represented by a Q function, is updated through dynamic programming to minimize the temporal difference (TD) error $L_{TD}(\phi)$, as shown by Eq. \ref{equ1}. 
\begin{equation}
\begin{aligned} \label{equ1}
L_{TD}(\phi)=\mathbb{E}_{({s}_{t},{a}_{t},{s}_{t+1})\sim\mathcal{D},{a}_{t+1}\sim\pi_{\theta^{\prime}}}\left[\left|\left|\left(r(s_{t},a_{t})+\gamma\min_{i=1,2}Q_{\phi_{i}^{\prime}}(s_{t+1},{a}_{t+1})\right)-Q_{\phi_{i}}(s_{t},{a}_{t})\right|\right|^{2}\right],
\end{aligned}
\end{equation}
where $Q_{\phi1}$ and $Q_{\phi1}$ are the main Q-networks, and $Q_{\phi^{\prime1}}$ and $Q_{\phi^{\prime2}}$ are the target Q-networks \cite{hasselt2010double}. In the policy improvement process, our method uses regularization for value function penalization and policy exploration constraint. Let ${R_e}$ denote the regularization term. We rewrite the policy objective $J_\theta$ as Eq. \ref{equ2}.
\begin{equation}
\begin{aligned} \label{equ2}
J_{\theta} = \mathbb{E}_{({s},{a})\sim\mathcal{D} }[Q^{\pi_{\theta}}(s,a)] + {R_e},
\end{aligned}
\end{equation}
where $Q^{\pi_{\theta}}(s,a)$ represents Q-values under the policy $\pi_{\theta}$. 

\textbf{Diffusion models.} Diffusion models \cite{ho2020denoising,sohl2015deep,song2019generative} are the latent variable generative models with $p_\theta({x}_0):=\int p_\theta({x}_{0:T}) d{x}_{1:T}$, where ${x}_{1},...,{x}_{T}$ are latent variables with the same dimensionality as the data ${x}_{0}\sim p({x}_0)$. Diffusion models fix the approximate posterior $q(x_{1:T}|x_{0})$, known as the forward process or diffusion process, to a Markov chain that gradually adds Gaussian noise to the data in $T$ steps with a variance schedule $\beta_i \in \left \{ \beta_1,...,\beta_T \right \}$, as shown by Eq. \ref{equ3}.
\begin{equation}
\begin{aligned} \label{equ3}
\begin{split}
q({x}_{1:T}|{x}_0):=\prod_{t=1}^Tq({x}_t|{x}_{t-1}), \\
q({x}_t|{x}_{t-1}):=\mathcal{N}({x}_t;\sqrt{1-\beta_t}{x}_{t-1},\beta_t\bm{I}).
\end{split}
\end{aligned}
\end{equation}

The joint distribution $p_{\theta}({x}_{0:T})$ is commonly referred to as the \textit{reverse process}, which is defined as a Markov chain with learned Gaussian transitions starting at $p({x}_{T})=\mathcal{N}({x}_{T};\bm{0},\bm{I})$, as shown by Eq. \ref{equ4}.
\begin{equation}
\begin{aligned} \label{equ4}
\begin{split}
p_{\theta}({x}_{0:T}):=p({x}_{T})\prod_{t=1}^{T}p_{\theta}({x}_{t-1}|{x}_{t}), \\ 
p_{\theta}({x}_{t-1}|{x}_{t}):={\mathcal{N}}({x}_{t-1};\mu_{\theta}({x}_{t},t),\Sigma_{\theta}({x}_{t},t)),
\end{split}
\end{aligned}
\end{equation}

where $\mu_{\theta}$ and $\Sigma_{\theta}$  represent the predicted mean and variance, respectively. The training objective of the \textit{reverse process} is to maximize the evidence lower bound defined as $\mathbb{E}_q[\ln\frac{p_\theta({x}_{0:{T}})}{q({x}_{1:{T}}\mid{x}_0)}]$. Based on the Denoising Diffusion Probabilistic Model (DDPM) \cite{ho2020denoising}, we approximate the Evidence Lower Bound (ELBO) by using a simplified surrogate loss function $L_{d}(\theta)$, as shown by Eq. \ref{equ5}.

\begin{equation}
\begin{aligned} \label{equ5}
L_{d}(\theta)=\mathbb{E}_{i\sim[1,T],\epsilon\sim\mathcal{N}(\bm{0},\bm{I}),{x}_{0}\sim q}\Big[||\epsilon-\epsilon_{\theta}(x_{i},i)||^{2}\Big],
\end{aligned}
\end{equation}
where $\epsilon \sim \mathcal{N}(\bm{0},\bm{I})$ represents the Gaussian noise added in the forward process, and $\epsilon_{\theta}$ denotes the parametric model. After training, the sampling procedure gets sample $x\sim p(x_T)$ and then the reverse diffusion is used to get resampled at each step.

\textbf{Generative adversarial networks.} Generative adversarial networks \cite{goodfellow2014generative} model data distributions through an adversarial game between a generator $G$ and a discriminator $D$. The training objective $V(G,D)$ is denoted as a maximization and minimization (minimax) game between $G$ and $D$, as shown by Eq. \ref{equ6}.

\begin{equation}
\begin{aligned} \label{equ6}
V(G,D)=\min_G\max_D\mathbb{E}_{x\sim p_D}[\log(D(x))]+\mathbb{E}_{z\sim p(z)}[\log(1-D(G(z)))],
\end{aligned}
\end{equation}
where $D$ is trained to maximize the probability of correctly classifying real data and fake ones from $G$. For a fixed $G$, the optimal discriminator is $D_{G}^{*}(x)=\frac{p_{{X}}(x)}{p_{{X}}(x)+p_{G}(x)}$, where $p_{X}$, $p_{G}$, and $x$ denote the distribution of real data, the distribution of generated data, and the samples from these distributions, respectively. In this case, the minimax problem can be converted into an optimisation problem for the generator $G$: $\min_{G}JS(P_{X}(\cdot)||P_{G}(\cdot))$, where $JS(\cdot)$ denotes the Jensen-Shannon divergence \cite{menendez1997jensen}. The essence of the formula is that $G$ is trained to generate data that match the distribution of real data.

In this work, we use two different types of GANs in policy learning, Vanilla GAN \cite{goodfellow2014generative} and Wasserstain GAN (WGAN) \cite{arjovsky2017wasserstein}.
The objectives of vanilla GAN and WGAN are $\min_G\max_D\mathbb{E}_{x\sim p_D}[\log(D(x))]+\mathbb{E}_{z\sim p(z)}[\log(1-D(G(z)))]$ and  $\min_G\max_{D\in\mathcal{D}}\mathbb{E}_{x\sim p_{data}}\left[D(x)\right]-\mathbb{E}_{z\sim p_z}\left[D(G(z))\right]$, respectively, where ${\mathcal{D}}$ in WGAN denotes the set of $K-Lipschitz$ functions.

\section{Diffusion Policies with Generative Adversarial Networks}

\subsection{DiffPoGAN framework and diffusion policy}\label{sec4.1}

Motivated by the similarity of the actor-critic and GAN frameworks, we propose to utilize the generator in the GAN as the actor for RL. To enable the generator to maximize Q-values in RL while generating data that are realistic enough to fool the discriminator, and also to avoid the impact of mixed data distribution \cite{vuong2022dasco}, we use the diffusion as the policy generator due to its powerful generative capability to improve the expressiveness of learned polices, and the learned polices can be easily regularized to close behavior policies and shaped to visit in distribution (ID) data.

We consider the objective as shown by \ref{equflow}. The first term is the min-max optimization objective of GANs, and the second term is the maximum likelihood estimation between the generated data and the real data. The second term also serves as the secondary objective for the generator to optimize the distribution match between the generated policies and behavior policies. This objective  shown by Eq. \ref{equflow} is similar to the optimization objective in flow-GAN \cite{grover2018flow}, and the previous research \cite{zhang2024constrained} has shown that applying flow-GAN in offline RL can make an accurate modeling for state-action pairs.

\begin{equation}
\begin{aligned} \label{equflow}
V(G_\theta,D_\phi) = \min_{\theta}\max_{\phi}\mathcal{L}_{\mathcal{G}}(G_{\theta},D_{\phi})-\lambda\mathbb{E}_{\tau\sim P_{data}}[\log(p_{\theta}(\tau))],
\end{aligned}
\end{equation}
where $\tau$ is a trajectory which consists of $(s,a)$ pairs. $G_\theta$ and $D_\phi$ denote the networks of the generator and the discriminator, respectively.

We draw inspiration from the flow-GAN \cite{grover2018flow} and propose to incorporate MLE as a secondary objective. This helps the generator optimize the distribution match between the learned policies and behavior policies for maximizing the expected returns in the policy optimization process while generating actions that can fool the discriminator. However, fooling the discriminator requires more realistic and diverse data, so we replace the generator in the GAN with a diffusion model to generate expressive policies. In addition, we add the secondary objective MLE term to the optimization objective of the diffusion policy network and propose the objective as shown by Eq. \ref{equ9}.

\begin{equation}
\begin{aligned} \label{equ9}
\pi=\operatorname*{arg max}_{\pi}[\mathbb{E}_{(s,a)\sim\mathcal{D}}[Q(s,\pi_{\theta}(s))]+\lambda\mathbb{E}_{s\sim D,a\sim\pi_{\beta}(\cdot|s)}[\log(\pi_{\theta}(s,a))]],
\end{aligned}
\end{equation}
where $\lambda$ is a hyperparameter, and $\mathbb{E}_{s\sim D,a\sim\pi_{\beta}(\cdot|s)}[\log(\pi_{\theta}(\cdot|s))]$ denotes the MLE term. We main focus on the likelihood of the behavior policy $\pi_\beta(s,a)$ and the learned policy $\pi_\theta(s,a)$. We represent our policy $\pi_\theta$ in terms of a conditional diffusion, as shown by Eq. \ref{equ10}.

\begin{equation}
\begin{aligned} \label{equ10}
\pi_\theta({a}\mid s)=p_\theta({a}^{0:N}\mid s)=\mathcal{N}({a}^N;\bm{0},\bm{I})\prod_{i=1}^Np_\theta({a}^{i-1}\mid{a}^i,s),
\end{aligned}
\end{equation}

where the superscript denotes the time step in the diffusion process, and $a^0$ represents the end sample of the reverse chain and is used in the evaluation process of RL. $p_{\theta}(a^{0:N}\mid s)$ denotes the joint distribution of all noisy samples. According to DDPM \cite{ho2020denoising}, $p_{\theta}(a^{0:N}\mid s)$ can be modeled as a Gaussian distribution $\mathcal{N}({a}^{i-1};{\mu}_{\theta}({a}^{i},s,i),{\Sigma}_{\theta}({a}^{i},{s},i))$. We fix the covariance matrix as $\Sigma_{\theta}({a}^{i},{s},i)=\beta_{i}{I}$ and predict the mean of the noisy through the conditional noise model $\epsilon_\theta({a}^i,{s},i)$, as shown by Eq. \ref{equ11}.

\begin{equation}
\begin{aligned} \label{equ11}
\mu_\theta(a^i,s,i)=\frac{1}{\sqrt{\alpha_i}}\big(a^i-\frac{\beta_i}{\sqrt{1-\alpha_i}}\epsilon_\theta(a^i,s,i)\big).
\end{aligned}
\end{equation}

In the sampling process, we first sample $a^{N}\sim\mathcal{N}(\bm{0},\bm{I})$ and then sample from the reverse diffusion chain parameterized by $\theta$, as shown by Eq. \ref{equ12}.

\begin{equation}
\begin{aligned} \label{equ12}
a^{i-1}\mid a^{i}=\frac{a^{i}}{\sqrt{\alpha_{i}}}-\frac{\beta_{i}}{\sqrt{\alpha_{i}(1-\bar{\alpha}_{i})}}\epsilon_{\theta}(a^{i},s,i)+\sqrt{\beta_{i}}\epsilon, \ \ \epsilon\sim\mathcal{N}(\bm{0},\bm{I}), \ \ \mathrm{for}\  i=N,\ldots,1.
\end{aligned}
\end{equation}

We apply the same noise schedule as \cite{wang2022diffusion}, as shown by Eq. \ref{equ13}.

\begin{equation}
\begin{aligned} \label{equ13}
\beta_i=1-\alpha_i=1-e^{-\beta_{\min}(\frac{1}{T})-0.5(\beta_{\max}-\beta_{\min})\frac{2i-1}{T^2}},
\end{aligned}
\end{equation}
which is a noise schedule obtained under the variance preserving stochastic differential equation (SDE) of \cite{song2020score}.

Since behavior policies are hard to estimate, the computation of the MLE term in Eq. \ref{equ9} is difficult in practice. We prove that the MLE term can be approximated via the diffusion loss, which is shown in Appendix \ref{proof1}.

\textbf{Proposition 4.1.}\label{pro4.1} \textit{In the diffusion policies, the MLE between the learned policy distribution and behavior policy distribution can be approximated by MSE-like loss based on ELBO.}  
\begin{equation}
\begin{aligned} \label{equ14}
\mathbb{E}_{s\sim D,a\sim\pi_{\beta}(\cdot|s)}[\log(\pi_{\theta}(s,a))] \approx - L_d(\pi_\beta,\pi_\theta).
\end{aligned}
\end{equation}
where $L_d(\pi_\beta,\pi_\theta)$ is the diffusion loss.

According to the Proposition \ref{pro4.1}, the diffusion policies can be converted into the parametrized policies with the optimization objective, as shown by Eq. \ref{equ15}.

\begin{equation}
\begin{aligned} \label{equ15}
\pi=\arg\max_\pi\mathbb{E}_{s\sim\mathcal{D},a\sim\pi_\theta(a|s)}\bigg[Q(s,a)-\lambda L_d(\pi_\beta,\pi_\theta)\bigg]
\end{aligned}
\end{equation}
where $\lambda$ is a hyperparameter. In this case, we convert the MLE into a solvable parameterized network and satisfy the requirement of distribution match, which allow to generate sufficient data for fooling the discriminator.

\subsection{Regularization by the GAN}\label{sec4.2}

In Section \ref{sec4.1}, we develop a part of the optimization objective for generator, where the diffusion objective is introduced in RL. However, a simple policy generation scheme as shown in Eq. \ref {equ15} still suffers from the OOD actions because the actions of the diffusion policy derive from sampling operation, which can not guarantee that the actions are in distribution, and the diffusion loss can not accurately measure the impact of the sampled action. In this case, it is necessary to evaluate the authenticity of the sampled actions and perform a reasonable regularization. In our method, it is achieved through the discriminator in the GAN. In the process of maximizing the Q values, we propose an adaptive weight based on the discriminator output to impose a reasonable and effective constraint on the diffusion policy. 

For the posterior term $L_d(\pi_\beta,\pi_\theta)$ in Eq. \ref{equ15}, a lower $L_d$ indicates a greater log-likelihood between the diffusion policy distribution and behavior policy distribution. However, the actions engaged in policy evaluation are derived from sampling and not necessarily in distribution. To impose a reasonable regularization on diffusion policies and mitigate the uncertainty effect from sampled actions, we design a down-weight term $\frac{D(s,a)}{D(s,a_{Da}(s))}$, where $D$ denotes the discriminator, according to the authenticity of the sampled data. Besides, our method uses the probability assigned to the sampled actions to regularize the estimation of the value function in the policy objective, as shown by Eq. \ref{pinoal}.

\begin{equation}
\begin{aligned} \label{pinoal}
\pi=\arg\max_{\pi}\mathbb{E}_{s,a_{Da}\sim\mathcal{D},a\sim\pi_\theta(a|s)}\biggl[Q(s,a)-\lambda\frac{D(s,a)}{D(s,a_{Da}(s))}L_{d}(\pi_{\beta},\pi_\theta) + S_D\biggr],
\end{aligned}
\end{equation}
where $S_D$ denotes the evaluation of discriminator, which takes two types including $-log(D(s,a))$ and $log(1-D(s,a))$ in this paper.

\subsection{Updating method and practical algorithm}\label{sec4.3}

We incorporate the whole diffusion policy and GAN to train the policies for offline RL. The whole updating process are shown by Eqs. \ref{equ16} and \ref{equ17}.

\begin{equation}
\begin{aligned} \label{equ16}
Q^{k+1}\leftarrow\arg\min_{Q}\mathbb{E}_{({s}_{t},{a}_{t},{s}_{t+1})\sim\mathcal{D},{a}_{t+1}^{0}\sim\pi^k_{\theta^{\prime}}}\left[\left|\left|\left(r(s_{t},a_{t})+\gamma\min_{i=1,2}Q_{\phi_{i}^{\prime}}(s_{t+1},{a}_{t+1}^{0})\right)-Q_{\phi_{i}}(s_{t},{a}_{t})\right|\right|^{2}\right]
\end{aligned}
\end{equation}

\begin{equation}
\begin{aligned} \label{equ17}
\pi^{k+1}\leftarrow\arg\max_{\pi}\mathbb{E}_{s,a_{Da}\sim\mathcal{D},a\sim\pi^{k}_\theta(a|s)}\biggl[\alpha \cdot Q^{k+1}(s,a)-\lambda\frac{D^{k}(s,a)}{D^{k}(s,a_{Da}(s))}L_{d}(\pi_{\beta},\pi_\theta^{k}) + S_D\biggr],
\end{aligned}
\end{equation}
where $a_{Da}$ is the action from behavior policies, and $k$ denotes the $k^{th}$ step of the policy iteration. $D^{k}(s,a)$ is the score of the discriminator for the sampled actions, and $S_D$ is the discriminator's evaluation. $\alpha=\frac{\eta}{\mathbb{E}_{(s,{a})\sim\mathcal{D}}[|Q_{\phi}({s},{a})|]}$ is a normalization term based on the minibatch $\left \{ s,a \right \} $ with a pyperparameter $\eta$ according to \cite{fujimoto2021minimalist}. It should be noted that gradients are not propagated into the down-weight term.

After introducing the optimization objectives of the offline RL network, we define the update rules for the discriminator. There are two kinds of discriminators for the training process, whose optimization objectives $Loss_D$ are shown by Eqs. \ref{equ18} and \ref{equ19}, respectively.

\begin{equation}
\begin{aligned} \label{equ18}
Loss_D = -\mathbb{E}_{(s,a_{Da})\sim \mathcal{D}}[\log D(s,a_{Da})]-\mathbb{E}_{s\sim \mathcal{D},a\sim\pi_\theta(\cdot|s)}[\log(1-D(s,a))],
\end{aligned}
\end{equation}

\begin{equation}
\begin{aligned} \label{equ19}
Loss_D = \mathbb{E}_{s\sim \mathcal{D},a\sim\pi_\theta(\cdot|s)}[f_w(s,a)]-\mathbb{E}_{(s,a_{Da})\sim \mathcal{D}}[f_w(s,a_{Da})],
\end{aligned}
\end{equation}
where $f_w$ denotes the $Wasserstein$ distance. The two objective functions shown by Eqs. \ref{equ18} and \ref{equ19} are used in different tasks in our experiments, as shown in subSection 5.1.

Algorithm \ref{alg1} provides the overall procedure of our algorithm. At each training step, we sample a batch of training data from the offline dataset and then update the parameters of the value function, policy, and discriminator, respectively.

\begin{algorithm}[h]
	\caption{DiffPoGAN algorithm}
	\begin{algorithmic}[1]
		\STATE Initialize the policy network $\pi_{\theta}$, critic network $Q_{\phi1}$, $Q_{\phi2}$ and target network $\pi_{\theta'}$, $Q'_{\phi1}$, $Q'_{\phi2}$, discriminate $D_{\omega}$ and offline replay buffer $D$.
		\FOR{step $i = 0 $ to $T$}
		\STATE Sample minibatch of transitions $(s,a,r,s')$ from $D$
		\STATE Sample $a' \sim \pi_{\theta}(\cdot | s')$ according to Eq. \ref{equ12}
		\STATE $//$ \textbf{Discriminate updating}
		\STATE $\omega^{k+1}\leftarrow$  Update discriminator $D_\omega$ according to Eq. \ref{equ18} or Eq. \ref{equ19}
		\STATE $//$ \textbf{Q-value function updating}
		\STATE $\phi^{k+1}\leftarrow$ Update Q-function $Q_{\phi}$ using the Bellman update according to Eq. \ref{equ16}
            \IF{$i$ mod $d $}
    		\STATE $//$ \textbf{Policy updating}
    		\STATE Sample $a_{t}^{0} \sim \pi_{\theta(a_{t}|s_{t})}$ according to Eq. \ref{equ12}
    		\STATE $\theta^{k+1}\leftarrow$ Update policy $\pi_{\theta}$ according to Eq. \ref{equ17}
            \ENDIF
		\STATE $//$ \textbf{Target networks updating}
		\STATE $\theta^{\prime}=\rho\theta^{\prime}+(1-\rho)\theta$, $\phi_{i}^{\prime}=\rho\phi_{i}^{\prime}+(1-\rho)\phi_{i}$ for $i={1,2}$
		

		\ENDFOR
		
	\end{algorithmic}
	\label{alg1}
\end{algorithm}

\section{Experiments}

\subsection{Benchmarks and baselines}
We evaluate our method on the D4RL \cite{fu2020d4rl} benchmark with two different domains: Gym-MuJoCo and Antmaze. Gym-MuJoCo is a classical domain for evaluating locomotion tasks, where the data collected by the behavior policy only cover a small part of the state-action space, so it is appropriate to evaluate the methods of handling the OOD issue  \cite{fujimoto2019off,kumar2020conservative}. In the Gym-MuJoCo domain, we focus on three locomotion tasks: Walker2d-v2, Hopper-v2, and HalfCheetah-v2. All the three tasks contain three different datasets including medium, medium-expert, and medium-replay. The medium dataset of those tasks consists of data collected from a single sub-optimal policy, while the medium-expert dataset is generated by a mixture of data collected from sub-optimal and near-optimal policies. The medium-replay dataset contains the first million data collected from the soft actor-critic (SAC) algorithm \cite{haarnoja2018soft}. Antmaze is a domain of evaluating control and navigation tasks, where the data consist of undirected trajectories, and it is appropriate to evaluate the data reorganization capacity for RL methods. In the Antmaze domain, different tasks represent the mazes with different sizes and different complexity levels. 

The objective function in Eq. \ref{equ18} is mainly used in our experiments. Specially, the rewards in the Antmaze domain is modified according to CQL \cite{kumar2020conservative} in our method, which influences the scoring of the discriminator and makes the results negative and unable to be logarithmic. Hence, we select the vanilla GAN to calculate $-log(1-D)$, shown by Eq. \ref{equ19}, as the scoring method.

To demonstrate the performance of our DiffPoGAN, we compare it with five SOTA diffusion-based offline RL methods including Diffusion Q-learning \cite{wang2022diffusion}, IDQL \cite{hansen2023idql}, SFBC \cite{chen2022offline}, Diffuser \cite{janner2022planning}, and DiffCPS \cite{he2023diffcps}. Moreover, we also select IQL \cite{kostrikov2021offline} and CQL \cite{kumar2020conservative} as the comparative methods because of their competitive performance in offline RL. Implementation details are provided in Appendix \ref{expde}.

\begin{table*}
	\centering
 	\caption{Performance of DiffPoGAN and other SOTA methods on D4RL. The mean and standard deviation of the normalized scores for DiffPoGAN are averaged over 5 random seeds. We report the performance of baseline methods using the best results reported from their paper. "-A" refers to any number of hyperparameters allowed. The highest score for each experiment is bolded. med = medium, r = replay, e = expert, hc = halfcheetah, wa = walker2d, ho = hopper. u = umaze, med = medium, l = large, d = diverse, p = play.}
	\begin{tabularx}{\textwidth}{l@{\hspace{5pt}}c@{\hspace{5pt}}c@{\hspace{4pt}}c@{\hspace{4pt}}c@{\hspace{4pt}}c@{\hspace{4pt}}c@{\hspace{4pt}}c@{\hspace{4pt}}|@{\hspace{4pt}}c}
		\toprule\noalign{\smallskip}	
		Dataset & IQL & CQL & IDQL-A & SFBC & Diffuser  & Diffuison-QL & DiffCPS & DiffPoGAN(Ours)\\
		\midrule\noalign{\smallskip}	
		
		Hc-med & 47.4 & 44.4 & 51.0 & 	45.9 & 	44.2 & 51.1	& 71.0 & $\mathbf{71.2} \pm 0.9$\\
		Ho-med		 & 66.2 & 58.0 & 65.4 & 	57.1 & 	58.5 & 90.5 	& 100.1 & $\mathbf{101.3} \pm 2.1$\\
		Wa-med	 & 78.3 & 79.2 & 82.5 &	77.9 	 &   79.7 & 87.0 & $\mathbf{90.9}$ & $88.7 \pm 1.8$ \\
		
		Hc-med-exp & 86.7 & 62.4 & 95.9 	& 	92.6  & 	79.2 & 96.8 & 100.3 & $\mathbf{101.1\pm1.8}$ \\
		Ho-med-exp 		& 91.5 & 98.7 & 108.6 	& 	108.6 & 	107.2 &111.1 & $\mathbf{112.1}$ & $111.1 \pm 0.4 $\\
		Wa-med-exp 	& 109.6 & 110.1 & 112.7 & 	109.8 & 	108.4 &110.1 & $\mathbf{113.1}$ & $110.3 \pm 1.7$\\

		Hc-med-rep & 44.6 & 46.2 & 45.9 & 	37.1 & 	42.4 & 47.8 & 50.5 & $\mathbf{61.2 \pm 0.6}$ \\
		Ho-med-rep 		& 94.7 & 48.6 & 92.1 & 	86.2 & 	96.8 &101.3  & 101.1 & $\mathbf{102.1\pm 0.6}$\\

		Wa-med-rep 	& 73.8 & 26.7 & 85.1 & 	65.1 & 	61.2 &95.5 & 91.3 & $\mathbf{99.5\pm 1.4}$ \\

		\bottomrule \noalign{\smallskip}
  
		{Muj-mean} & 76.9 & 63.9 & 82.1 & 75.6 & 75.3 & 87.9 & 92.26 & $\bm{94.06}$\\
		\bottomrule	\noalign{\smallskip}
		
		Ant-u  & 87.5 & 74.0 & 94.0 & 92.0 & - & 93.4 & 97.4 &$ \bm{99.4\pm 0.8}$ \\
		Ant-u-d  & 62.2 & 84.0 & 80.2 & 85.3 & - & 66.2 & 87.4 &$\bm{91.0\pm 3.3}$ \\
		Ant-m-p & 71.2 & 61.2 & 84.5 & 81.3 & - & 76.6 & $\bm{88.2}$ &$85.3\pm 5.1$ \\
		Ant-m-d & 70.0 & 53.7 & 84.8 & 82.0 & - & 78.6 & 87.8 &$\bm{89.3\pm 3.2}$ \\
		Ant-l-p  & 39.6 & 15.8 & 63.5 & 59.3 & - & 46.4 &$\bm{65.6}$ &$60.4\pm 4.7$\\
		Ant-l-d  & 47.5 & 14.9 & 67.9 & 45.5 & - & 57.3 & 63.6 &$\bm{69.8\pm 2.2}$ \\
		\bottomrule \noalign{\smallskip}
		{Ant-mean} & 63.0 & 50.6 & 79.1 & 74.2 & - & 69.8 & 81.67 & $\bm{82.5}$\\	\noalign{\smallskip}
		\bottomrule
	\end{tabularx}

	\label{tab1}
\end{table*}

\subsection{Comparative results with SOTA methods}\label{comp}

The data in Table \ref{tab1} show that DiffPoGAN achieves competitive results compared with other offline RL methods. DiffPoGAN produces the best average results in both Gym-MuJoCo and Antmaze domains, and outperforms the baseline method diffusion Q-learning by $7\%$ and $18\%$, respectively, which is owing to highly expressive policies and the effective regularization method in DiffPoGAN. Especially in the medium-replay dataset that contains a lot of bad data from the early training stages and makes offline RL algorithms arduous to learn optimal policies, DiffPoGAN achieves the best performance among all the comparative methods. Highly expressive policies can free DiffPoGAN from the constraints of bad data, and are combined with the proposed regularization method to learn good policies. In the Antmaze domain with sparse rewards and the large amount of undirected trajectories, the proposed DiffPoGAN can also achieve competitive results. In some small-scenario tasks like umaze, DiffPoGAN achieves $100\pm 0\%$ success rate on some seeds, because the scoring of the sampled actions by the discriminator can help the Q function make a reasonable estimation of the current action, which effectively avoids the agent from selecting bad data and ensures a stable learning process. In the medium-play and large-play tasks, our method performs a little worse than DiffCPS because the double regularization in DiffPoGAN reduces the sensitivity to a single behavior policy, which leads to performance degradation in some special tasks.

\subsection{Ablation Study}
We conduct ablation study on two components that affect the performance of the DiffPoGAN method, including the number of diffusion step and the regularization term based on the discriminator output in the GAN.

\textbf{Diffusion Step:} The step number $T$ is an important hyperparameter in the diffusion since it affects the quality of the learned policy. We evaluate DiffPoGAN with different $T$ on Gym-MuJoCo domains, and the average results show that $T=5$ is the best value for training, as shown in Fig. \ref{fig1}. More details can be found in Appendix \ref{expde}. We notice that the performance gradually decreases as $T$ increases. Considering the characteristics of offline data, a large $T$\begin{wrapfigure}{r}{0.7\textwidth}
    \centering
    
 \renewcommand{\arraystretch}{1}
    \renewcommand{\tabcolsep}{1.0pt}
	\centering

	\includegraphics[width=7cm,height=5cm]{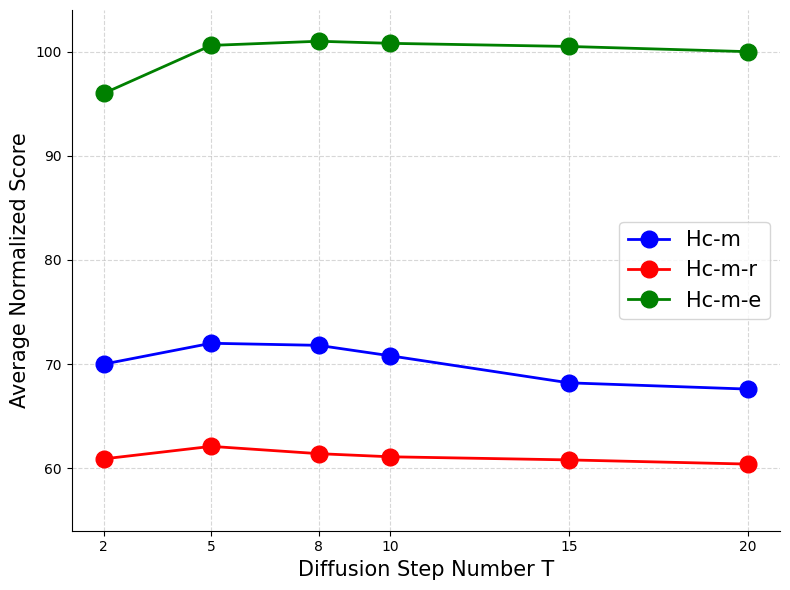}
 	\caption{Ablation studies of diffusion step number $T$ on the selected Gym tasks. We set $T$ in follow list for training: [2,5,8,10,15,20], and find that $T=5$ is better for training. HC = Halfcheetah, m = medium,r = replay,e = expert.}
  \vspace{-2mm}
	\label{fig1}
\end{wrapfigure} reduces the policy diversity, and then causes the policy to converge to a sub-optimal unimodal policy distribution. In contrast, a small $T$ can preserve the policy exploration capability, which can be effectively combined with the proposed DiffPoGAN method to make the learned policies converge to the optimal result. However, a too small $T$, e.g. $T=2$, means too few sampling steps and too weak network expression, which also leads to performance degradation. In conclusion, we select $T=5$ in our experiments on all compared benchmarks.

\textbf{Down-weight term based on the discriminator:} The down-weight term from the discriminator, which scores the authenticity of the sampling results, is crucial to obtain good performance in our method. To demonstrate the effectiveness of the down-weight term based on\begin{wraptable}{r}{0.5\textwidth}
    \centering
    
 \renewcommand{\arraystretch}{1}
    \renewcommand{\tabcolsep}{1.0pt}  
    \vspace{-2mm}
 \caption{Experiment for training without and with the down-weight term. The results are obtained from 3 random seeds.}
	\begin{tabular}{l|cc}
		\toprule\noalign{\smallskip}	
		Tasks & Without & With \\
		\midrule\noalign{\smallskip}	
            Halfcheetah-medium-replay & 57.6&$\bm{61.8}$\\
            Hopper-medium-replay & 99.2&$\bm{101.6}$\\
            Walker2d-medium-replay& 91.5&$\bm{98.6}$\\
            Antmaze-medium-play& 79&$\bm{85}$\\
            Antmaze-medium-diverse& 82&$\bm{88}$\\
            \midrule\noalign{\smallskip}
            average score&81.8&$\bm{87}$\\
        \bottomrule	
 \end{tabular}
 \label{tab3}
\end{wraptable} discriminator in the GAN, we conduct an additional experiment that evaluates the performances of the proposed DiffPoGAN with and without the down-weight term on the selected tasks. As shown in Table \ref{tab3}, the average score of the method with down-weights exceeds the method without it by $6.4\%$, which means that the proposed down-weight term provides a reasonable and conservative weakening scheme for the diffusion loss to avoid overestimation for the Q value. Besides, the improvement is more applicable for the Antmaze tasks because the down-weight term enables a more rational evaluation for actions, especially in the sparse reward environment in Antmaze.

\section{Conclusion}\label{conc}
In this paper, we propose the DiffPoGAN, an offline RL method combining the diffusion and GAN, to address the challenge of fooling the discriminator while maximizing the expected returns when using the diffusion as a policy generator. In DiffPoGAN, we introduce the GAN framework into offline RL, where the generator is served as the actor and the discriminator provides the probability that generated actions are in behavior policies. Besides, we develop an additional regularization term based on the discriminator output to appropriately constrain policy exploration for policy improvement. Experimental results on the D4RL benchmark demonstrate that DiffPoGAN outperforms SOTA methods. Future work will focus on how to reduce the effect from double regularization in DiffPoGAN and further improve the performance.

\bibliographystyle{unsrt}
\bibliography{cites}


\end{document}